\pdfoutput=1

\documentclass[conference,compsoc]{IEEEtran}
\IEEEoverridecommandlockouts
\makeatletter
\def\ps@IEEEtitlepagestyle{%
  \def\@oddfoot{\mycopyrightnotice}%
  \def\@evenfoot{}%
}
\def\mycopyrightnotice{%
  {\footnotesize The copyright belongs to me!\hfill}
  \gdef\mycopyrightnotice{}
}

\usepackage{url}            
\usepackage{booktabs}       
\usepackage{amsfonts}       
\usepackage{graphicx}
\graphicspath{{./figures/}}
\usepackage{amsmath}
\usepackage{algorithm}
\usepackage[]{algpseudocode}

\usepackage{tikzsymbols}

\usepackage{enumitem}

\usepackage{footnote}
\makesavenoteenv{tabular}
\makesavenoteenv{table}

\ifCLASSOPTIONcompsoc
  \usepackage[nocompress]{cite}
\else
  \usepackage{cite}
\fi

\ifCLASSINFOpdf
\else
\fi

\let\OLDthebibliography\thebibliography
\renewcommand\thebibliography[1]{
  \OLDthebibliography{#1}
  \setlength{\parskip}{-1pt}
  \setlength{\itemsep}{0pt plus 0.2ex}
}

\usepackage{tikz}
\newcommand*\circled[1]{\tikz[baseline=(char.base)]{
            \node[shape=circle,draw,inner sep=0.5pt] (char) {#1};}}

\begin{document}


%
\title{\vspace{-30pt}{\footnotesize {\normalfont This paper will appear
in the proceedings of DATE 2018. Pre-print version,
for personal use only.}}\\~\\
HyperPower: Power- and Memory-Constrained Hyper-Parameter\\
Optimization for Neural Networks}


\author{\IEEEauthorblockN{Dimitrios~Stamoulis\IEEEauthorrefmark{1}, Ermao~Cai\IEEEauthorrefmark{1},
Da-Cheng~Juan\IEEEauthorrefmark{2}, Diana~Marculescu\IEEEauthorrefmark{1}}
\IEEEauthorblockA{\IEEEauthorrefmark{1}Department of ECE, Carnegie Mellon University, Pittsburgh, PA\\
\IEEEauthorrefmark{2}Google Research, Mountain View, CA\\
Emails: dstamoul@andrew.cmu.edu, ermao@cmu.edu, dacheng@google.com, dianam@cmu.edu}}

\maketitle

\thispagestyle{plain}
\pagestyle{plain}


\begin{abstract}
While selecting the hyper-parameters of Neural Networks (NNs)
has been so far treated as \emph{an art}, the emergence
of more complex, deeper architectures poses increasingly
more challenges to designers and Machine Learning (ML) practitioners, especially
when power and memory constraints need to be considered.
In this work, we propose \emph{HyperPower}, a framework that enables efficient
Bayesian optimization and random search in the context of power- and
memory-constrained hyper-parameter optimization for NNs running on a given
hardware platform. \emph{HyperPower}
is the first work (i) to show that power consumption can be used
as a low-cost, \emph{a~priori} known constraint, and (ii) to propose
predictive models for the power and memory of NNs executing on GPUs.
Thanks to \emph{HyperPower}, the number of function evaluations
and the best test error achieved by a constraint-unaware method are
reached up to $\textbf{112.99}\times$ and $\textbf{30.12}\times$ faster, respectively,
while \textbf{never} considering invalid configurations.
\emph{HyperPower} significantly speeds up the hyper-parameter optimization,
achieving up to $\textbf{57.20}\times$ more function evaluations compared to
constraint-unaware methods for a given time interval, effectively
yielding significant accuracy improvements by up to $\textbf{67.6\%}$.
\end{abstract}


%
\IEEEpeerreviewmaketitle




\section{Introduction}

Hyper-parameter optimization of Machine Learning algorithms, and especially of Neural Networks (NNs),
has emerged as an increasingly challenging and expensive process, dubbed by many
researchers to be \emph{more of an art than science}. A surprisingly high number of
state-of-the-art methodologies heavily relies
on \emph{human experts} and this knowledge separates
a useless model from cutting edge performance~\cite{shahriari2016taking}.
Nonetheless, as the design space of hyper-parameters
to be tuned grows, the task of proper hand-tailored tuning can become daunting~\cite{swersky2013multi}.

\begin{figure}[ht!]
  \centering
  \vspace{-3pt}
  \includegraphics[width=1.05\columnwidth]{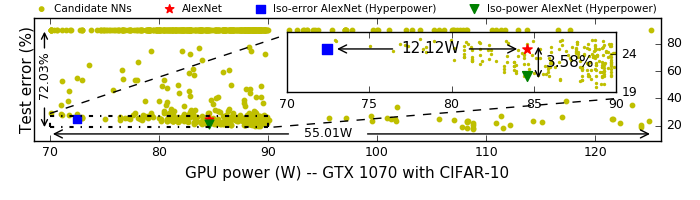}
  \vspace{-25pt}
  \caption{Power differs for a given accuracy level,
  hampering a human expert's ability to identify the optimal NN configuration
  under hardware constraints (similarly for memory, not shown
  due to space limitations).}
  \label{fig:motivation}
\end{figure}

Moreover, the ability of a \emph{human expert}
could be hampered further if we are to consider platform specific test-time power and memory constraints.
We visualize the complexity of the design space in Figure~\ref{fig:motivation},
by reporting testing error and GPU power consumption for different NN variations of the
AlexNet (CIFAR-10 with Caffe~\cite{jia2014caffe} on Nvidia GTX-1070).
We observe that, for a given accuracy level, power could differ
significantly by up to $55.01W$ (\emph{i.e.}, more than a third of the GPU Thermal Design
Power). Hence, the \emph{more of an art than science} rationale could become
non-trivial to exploit in a hardware constrained design space, necessitating a significant,
yet often unavailable, familiarity of the researcher with the hardware architecture.
In addition, the evaluation of each possible architecture could take hours, if not days, to train.
Thus, traditional techniques for hyper-parameter optimization, such as grid search, yield
poor results in terms of performance and training time~\cite{swersky2013multi}.

\emph{Bayesian optimization} and \emph{random search} methods
have been shown to outperform human experts in hyper-parameter optimization for
NNs~\cite{snoek2012practical}\cite{swersky2013multi}\cite{bergstra2012random}.
Nevertheless, both methods have not been extensively studied in the context
of hardware-constrained hyper-parameter optimization. On the one hand, Bayesian optimization
has increased complexity for cases where constraints are not available
\emph{a priori}~\cite{gelbart2014bayesian}. On the other hand, random methods
without hardware-aware  enhancements could wastefully consider infeasible points
that are randomly selected. These observations constitute
the key \textbf{motivation} behind our work.
As a motivating example, hardware-aware hyper-parameter optimization
(based on our framework presented later on) can find an iso-error NN with
power savings of $12.12W$ compared to AlexNet, or an iso-power NN with error decreased to $21.16$ from $24.74\%$.

\begin{figure*}
  \centering
  \includegraphics[width=7.1in, height=1.3in]{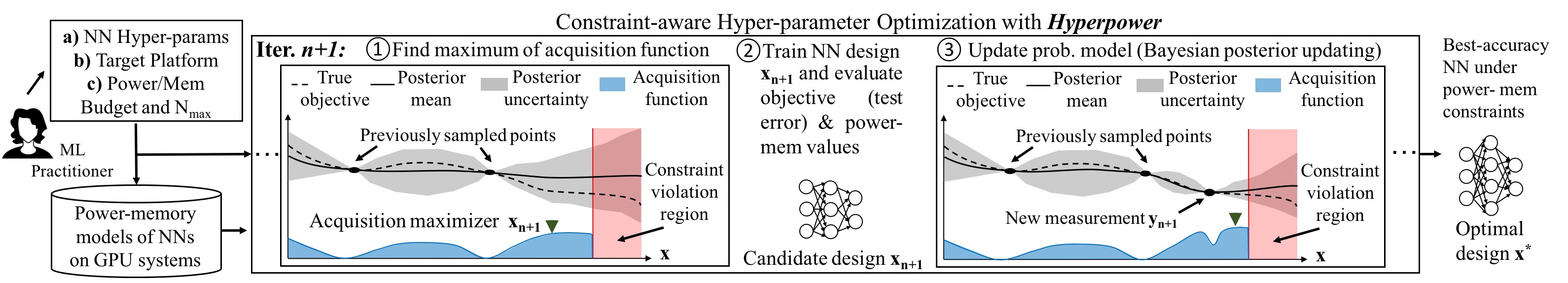}
  \vspace{-20pt}
  \caption{Overview of \emph{HyperPower} flow and illustration of the Bayesian optimization
  procedure during each iteration $n+1$. The ML designer only provides the NN design space,
  the target platform, the power/memory budget values, and the number of iterations $N_{\text{max}}$.
  Our goal is to find the NN configuration with minimum test error under
  hardware constraints. Bayesian optimization uses a surrogate probabilistic model $\mathcal{M}$ to
  approximate the objective function; the plots show the mean and confidence intervals estimated with the model
  (the true objective function is shown for reference, but it is unknown in practice). At each iteration
  $n+1$, an acquisition function $\alpha(\cdot)$ is expressed based on the model $\mathcal{M}$ and the maximizer of $\alpha(\cdot)$
  is selected as the candidate design point $\textbf{x}_{n+1}$ to evaluate. \emph{HyperPower} incorporates
  the power-memory models directly into the acquisition function formulation, thus inherently
  preventing sampling from constraint-violating regions (shaded red). The objective function (NN test error) is
  evaluated, \emph{i.e.}, the candidate NN design $\textbf{x}_{n+1}$ is trained and tested. Then,
  the probabilistic model $\mathcal{M}$ is refined via Bayesian posterior updating based on the new observation.
  After $N_{\text{max}}$ iterations, \emph{HyperPower} returns the design $\textbf{x}^*$ with
  optimal accuracy that satisfies the hardware constraints.}
  \vspace{-10pt}
  \label{fig:hyperpower}
\end{figure*}

To this end, we propose \emph{HyperPower}, a framework that enables efficient
Bayesian optimization and random search in the context of power- and memory-constrained
hyper-parameter optimization for NNs. Our work makes the
following \textbf{contributions}:
\begin{enumerate}[leftmargin=13pt]
\item \emph{HyperPower} is the \emph{first work} to show
that power consumption can be treated as an \emph{a priori} known constraint for NN model selection.
This insight for low-cost constraint evaluation
paves the way towards enabling efficient
Bayesian optimization and random search methods.
\item To the best of our knowledge, \emph{HyperPower} is the \emph{first} work to propose
predictive models for the power and memory consumption of NNs running on GPUs.
\item \emph{HyperPower} reaches the number of function evaluations
and the best test error of a constraint-unaware method up to $\textbf{112.99}\times$
and $\textbf{30.12}\times$ faster, respectively, while \textbf{never} considering invalid configurations.
\item \emph{HyperPower} allows for up to $\textbf{57.20}\times$ more function evaluations
compared to a constraint-unaware method for a given time interval,
yielding a significant accuracy improvement by up to $\textbf{67.6\%}$.
\end{enumerate}

\vspace{-5pt}
\section{Related work}
\label{sec:sota}

\vspace{-5pt}
\textbf{Modeling hardware metrics}:
Prior work relies on simplistic proxies of the memory consumption
(\emph{e.g.}, counts of the NN's weights~\cite{gelbart2014bayesian}), or on extrapolation
based on technology node energy tables per operation~\cite{rouhani2016energy}\cite{smithson2016opal}\cite{han2015learning}.
Consequently, existing modeling assumptions are either overly simplifying and have not
been compared against real platforms, or they reflect outdated technology nodes
that are not representative of modern GPU architectures. On the contrary,
we train our models on commercial Nvidia GPUs and we achieve accurate predictions 
against actual hardware measurements. Our recent work also introduces more elaborate (layer-wise)
predictive models for runtime and energy, which can be
incorporated into \emph{HyperPower}~\cite{cai2017neuralpower} and which could be flexibly extended to
account for process variations~\cite{stamoulis2016can},
thermal effects~\cite{cai2016exploring}, and aging~\cite{stamoulis2015efficient}. 

\textbf{Bayesian optimization under constraints}:
Prior art has proposed formulations for constrained Bayesian optimization
that generalize the model-based treatment of the objective
to the constraint functions~\cite{gelbart2014bayesian}.
Hern{\'a}ndez-Lobato \emph{et al.} have developed a general
framework for employing Bayesian optimization with unknown constraints or with
multiple objective terms~\cite{hernandez2016general}.
This framework has been successfully used for the co-design of hardware accelerators
and NNs~\cite{hernandez2016designing}~\cite{reagen2017case}, and the design of NNs under
runtime constraints~\cite{hernandez2016general}.
However, existing methodologies evaluate only MNIST on
hardware simulators~\cite{hernandez2016designing}~\cite{reagen2017case},
do not consider power as key design constraint~\cite{hernandez2016general},
and use a simplistic count of the network's weights as a proxy
for the memory constraint. Instead, in our work, we propose an accurate
model for both power and memory that is trained and tested on different
commercial GPUs and datasets.
\emph{HyperPower} and the proposed power and memory models can be flexibly
incorporated into generic formulations that support
constrained multi-objective optimization~\cite{hernandez2016general}.

Prior art has motivated optimization cases where the constraints can be expressed as
known \emph{a priori}~\cite{gelbart2014bayesian}; these formulations enable
models that can directly capture candidate configurations as valid or invalid~\cite{gramacy2010optimization}.
In this work, we are first to
show that both power and memory constraints can be formulated as constraints known \emph{a priori}.
We exploit this insight to train predictive models on the power and memory consumption of NNs
executing on state-of-the-art platforms and datasets,
allowing the \emph{HyperPower} framework to navigate the design space
in a constraint ``complying'' manner.
We extend the hyper-parameter optimization models to explicitly account for hardware imposed constraints.

\textbf{Random (model-free) methods}: Random~\cite{bergstra2012random}
and random-walk~\cite{smithson2016opal} hyper-parameter selection
has been shown to perform well in problems that have
low \emph{effective dimensionality}~\cite{shahriari2016taking}.
Nevertheless, when hardware constraints are
to be accounted for,  it is as likely for random methods to sample a point inside the
invalid region as to select a candidate point outside of it. 
Our enhancements allow  to quickly discard invalid randomly selected points.

\vspace{-5pt}
\section{HyperPower Framework}
\label{sec:method}

\vspace{-5pt}
The \emph{HyperPower} framework is illustrated in Figure~\ref{fig:hyperpower}.
The ML practitioner selects the hyper-parameter space (possible NN configurations), the target platform,
and the power/memory constraints. After $N_{\text{max}}$ iterations,
\emph{HyperPower} returns the NN with optimal accuracy that satisfies the hardware constraints.
This problem of interest is a special case of optimizing
function $f(\textbf{x})$ over design space $\mathcal{X}$ and constraints $\textbf{g}(\textbf{x})$,
\emph{i.e.}, ${\min}_{{\textbf{x} \in \mathcal{X}}} ~f(\textbf{x}), s.t. ~ \textbf{g}(\textbf{x}) \leq \textbf{c}$,
where the objective function (\emph{i.e.}, test error of each NN configuration) has no simple
closed form and its evaluations are costly. To efficiently solve this problem, \emph{HyperPower}
exploits the effectiveness of Bayesian optimization methods.

\vspace{-7pt}
\subsection{Bayesian optimization}
\vspace{-7pt}

Bayesian optimization is a sequential model-based approach that approximates the objective function
with a surrogate (cheaper to evaluate) probabilistic model $\mathcal{M}$, based on Gaussian processes (GP).
The GP model is a probability distribution over the possible functions of $f(\textbf{x})$, and it
approximates the objective at each iteration $n+1$ based on
data $\textbf{X}:= {x_i \in \mathcal{X}}_{i=1}^{n}$ queried so far.
We assume that the values $\textbf{f} := f_{1:n}$
of the objective function at points $\textbf{X}$ are jointly Gaussian
with mean $\textbf{m}$ and covariance $\textbf{K}$, \emph{i.e.},
$\textbf{f} ~| ~\textbf{X} \sim  \mathcal{N}(\textbf{m}, \textbf{K})$.
This formulation intuitively encapsulates our
belief about the shape of functions that are more likely to fit the data observed so far.
Since the observations $\textbf{f}$ are noisy
with additive noise $\epsilon \sim  \mathcal{N} (0,\sigma^2)$, we write the GP model as 
$\textbf{y} ~|~ \textbf{f}, \sigma^2 \sim \mathcal{N} (\textbf{f} , \sigma^2 \textbf{I})$.
At each point $\textbf{x} $, GP gives us a cheap approximation
for the mean and the uncertainty of the objective, written
as $p_\mathcal{M} (y|\textbf{x})$ and illustrated in Figure~\ref{fig:hyperpower} with the black curve and
the grey shaded areas.

Each iteration $n+1$ of a Bayesian optimization algorithm consists of three key steps:

\circled{1}~\textbf{Maximization of acquisition function}:
We first need to select the point $\textbf{x}_{n+1}$
(\emph{i.e.}, next candidate NN configuration) at which the objective (\emph{i.e.}, the test error of the candidate NN)
will be evaluated next. This task of guiding the search relies on the
so-called acquisition function $\alpha(\textbf{x})$. 
A popular choice for the acquisition function is the Expectation Improvement (EI) criterion,
which computes the probability that the objective function $f$ will exceed (negatively) some threshold $y^+$,
\emph{i.e.}, $EI(\textbf{x}) =  \int_{-\infty}^{\infty} \max\{y^+ - y, 0\} \cdot p_\mathcal{M} (y|\textbf{x}) ~ dy$.
Intuitively, $\alpha(\textbf{x})$ provides a measure of the direction toward
which there is an expectation of improvement of the objective function.

The acquisition function is evaluated at different candidate points $\textbf{x}$, yielding
high values at points where the GP's uncertainty is high (\emph{i.e.}, favoring exploration),
and where the GP predicts a high objective (\emph{i.e.}, favoring exploitation)~\cite{shahriari2016taking};
this is qualitatively illustrated in Figure~\ref{fig:hyperpower} (blue curve).
We select the maximizer of $\alpha(\textbf{x})$ as the point $\textbf{x}_{n+1}$ to evaluate next
(green triangle in Figure~\ref{fig:hyperpower}).
To enable power- and memory-aware Bayesian optimization, \emph{HyperPower} incorporates
hardware-awareness directly into the acquisition function (subsection~\ref{sub:ei}).

\circled{2}~\textbf{Evaluation of the objective}: Once
the current candidate NN design $\textbf{x}_{n+1}$ has been selected, the NN is generated
and trained to completion to acquire the test error. This is the most expensive step.
Hence, our efforts towards enabling efficient Bayesian optimization, mainly focus on detecting
when this step can be bypassed (subsection~\ref{sub:enh}).

\circled{3}~\textbf{Probabilistic model
update}: As the new objective value $y_{n+1}$ becomes available at the end
of iteration $n+1$, the probabilistic model
$p_\mathcal{M} (y)$ is refined via Bayesian posterior updating
(the posterior mean $\textbf{m}_{n+1} (\textbf{x})$ and covariance covariance $\textbf{K}_{n+1}$
can be analytically derived).
This step is quantitatively illustrated in Figure~\ref{fig:hyperpower} with the black curve and
the grey shaded areas. Please note how the updated model has reduced uncertainty around the previous samples and
newly observed point. 
For an overview of GP models 
the reader is referred to~\cite{shahriari2016taking}. 

\vspace{-7pt}
\subsection{\emph{HyperPower} enhancements}
\label{sub:enh}

\begin{figure}[t!]
  \centering
  \vspace{-5pt}
  \hspace*{-5pt}\includegraphics[width=0.52\columnwidth]{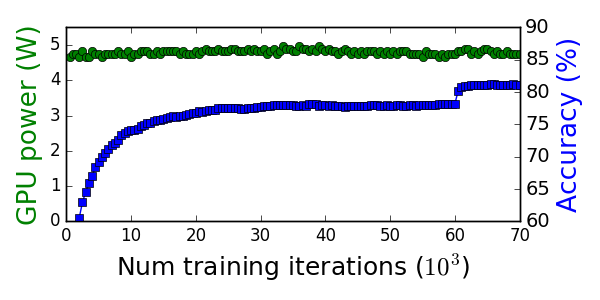}~
  \hspace*{-7pt}\includegraphics[width=0.52\columnwidth]{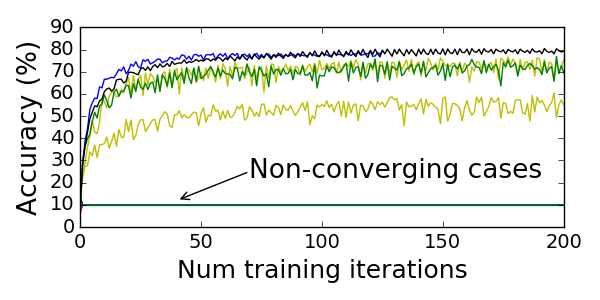}
  \vspace{-18pt}
  \caption{Visualizing our \textbf{insights}: how power varies vs accuracy with the
  number of training epochs (left); how accuracy can indicate configurations that do not converge
  to high-accuracy values ($>10\%$) (right).}
  \label{fig:insights}
  \vspace{-10pt}
\end{figure}

\vspace{-7pt}
\textbf{Early termination of the NN training at
step} \circled{2}: First, we observe that
candidate architectures that diverge during training can be \emph{quickly identified}
only after a few training epochs (Figure~\ref{fig:insights} (right)).
Please note that this is different than predicting the final test error
of a network, which could suffer from
overestimation issues~\cite{domhan2015speeding}, introducing artifacts to
the probabilistic model. Instead of predicting
for converging cases, we identify diverging cases, allowing the
optimization process to discard low-performance samples.

\textbf{Power and memory as low-cost constraints}:
To enable an efficient formulation with \emph{a-priori} known constraints,
we observe that power and memory are low-cost constraints to evaluate.
That is, as motivated in prior art for runtime~\cite{gelbart2014bayesian}~\cite{hernandez2016designing},
the power-memory characteristics of an NN
are not affected by the quality of the trained model itself.
In Figure~\ref{fig:insights} (left), we observe that the NN power values
on Nvidia TX1 with MNIST 
do not heavily change even if the NN is trained for more iterations
(unlike accuracy, obviously). We are \emph{first} to exploit this insight to
train predictive models for the power and memory of NN
architectures. More importantly, we use the predictive models
to formulate a power- and memory-constrained acquisition function.

\vspace{-9pt}
\subsection{Power and memory models}

\vspace{-9pt}
To enable \emph{a priori} power and memory constraint evaluations that are
decoupled from the expensive objective evaluation,
we propose to model power and memory consumption of an network as a function of
the $J$ discrete (structural) hyper-parameters $\textbf{z} \in \mathbb{Z}_{+}^J$ (subset of
$\textbf{x} \in \mathcal{X}$); we train on the structural hyper-parameters
$\textbf{z}$ 
that affect the NN's power and memory (\emph{e.g.}, number of hidden units),
since parameters such as learning rate have negligible impact.

To this end, we employ offline random sampling by generating
different configurations based on the ranges of the considered
hyper-parameters $\textbf{z}$ (discussed in Section~\ref{sec:setup}).
Since the Bayesian optimization corresponds to function evaluations with respect to the
test error\cite{snoek2012practical}, for each candidate design $\textbf{z}_l$
we measure the hardware platform's power $P_l$ and memory $M_l$ values during inference and
not during the NN's training. Given the $L$ profiled data points $\{(\textbf{z}_l,
P_l, M_l)\}_{l=1}^L$, we train the following models that are linear with respect to both
the input vector $\textbf{z} \in \mathbb{Z}_{+}^J$ and model weights $\textbf{w},\textbf{m}
\in \mathbb{R}^J$, \emph{i.e.}:
\vspace{-7pt}
\begin{equation}
\label{eq:pow}
\textbf{Power model}:~~\mathcal{P}(\textbf{z}) = \sum_{j=1}^{J} w_j \cdot z_j
\end{equation}
\vspace{-7pt}
\begin{equation}
\label{eq:mem}
\textbf{Memory model}:~~\mathcal{M}(\textbf{z}) = \sum_{j=1}^{J} m_j \cdot z_j
\end{equation}
We train the models above by employing a 10-fold cross validation on the dataset $\{(\textbf{z}_l,
P_l, M_l)\}_{l=1}^L$. While we experimented with nonlinear regression formulations which can be
plugged-in to the models (\emph{e.g.}, see our recent work~\cite{cai2017neuralpower}),
these linear functions provide sufficient accuracy (as shown
in Section~\ref{sec:results}). More importantly,
we select the linear form since it allows for the efficient evaluation of
the power and memory predictions within the acquisition function (next subsection), computed
on each sampled grid point of the hyper-parameter space. 

\begin{figure*}
  \centering
  \hspace*{-5pt}\includegraphics[width=0.70\columnwidth]{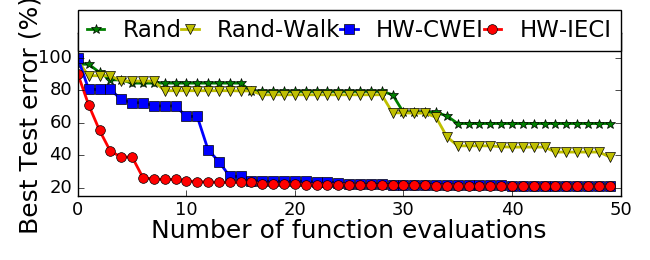}~
  \hspace*{-11pt}\includegraphics[width=0.70\columnwidth]{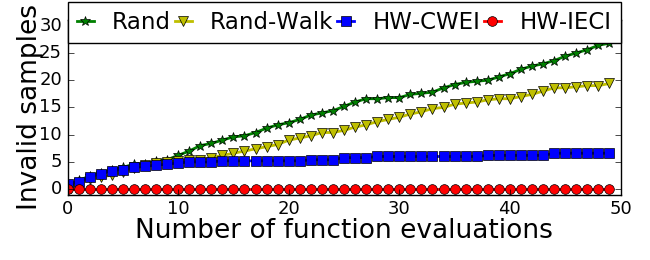}~
  \hspace*{-11pt}\includegraphics[width=0.70\columnwidth]{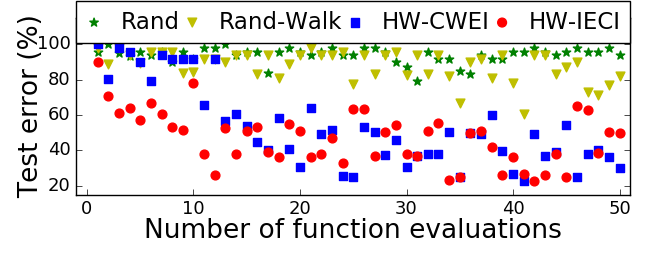}
  \vspace*{-12pt}
  \caption{Assessment of the four methods on hyper-parameter optimization on CIFAR-10 CNN.
  (left) Best observed test error against the number of function evaluations.
  (center) Number of constraint-violating samples against the number
  of function evaluations.  (right) Test error per function evaluation.
  }
  \label{fig:errors}
\end{figure*}

\vspace{-7pt}
\subsection{Constraint-aware acquisition function} 
\label{sub:ei}

\vspace{-7pt}
\textbf{\emph{HW-IECI}}: In the context of hardware-constraint optimization, EI allows us to
directly incorporate the \emph{a priori} constraint
information in a representative way. Inspired by constraint-aware 
heuristics~\cite{gelbart2014bayesian}~\cite{gramacy2010optimization},
we propose a power and memory constraint-aware acquisition function:
\begin{equation}
\label{eq:con_ei}
\begin{split}
a(\textbf{x}) =  &\int_{-\infty}^{\infty} \max\{y^+ - y, 0\} \cdot p_\mathcal{M} (y|\textbf{x}) \cdot \\
&  \mathbb{I}[\mathcal{P}(\textbf{z}) \leq \text{PB}] \cdot \mathbb{I}[\mathcal{M}(\textbf{z}) \leq \text{MB}] ~ dy
\end{split}
\end{equation}
where $\textbf{z}$ are the structural hyper-parameters, 
$p_\mathcal{M} (y|\textbf{x})$ is the predictive marginal density
of the objective function at $\textbf{x}$ based on surrogate model $M$.
$\mathbb{I}[\mathcal{P}(\textbf{z}) \leq \text{PB}]$ and $\mathbb{I}[\mathcal{M}(\textbf{z}) \leq
\text{MB}]$ are the indicator functions, which are equal to $1$ if
the power budget PB and the memory budget MB are respectively satisfied.
Typically, the threshold $y^+$ is adaptively set to the best value
$y^+ = \max_{i=1:n} y_i$ over previous observations~\cite{shahriari2016taking}\cite{gelbart2014bayesian}.

Intuitively, we capture the fact that improvement should not
be possible in regions where the constraints are violated.
Inspired by the integrated expected conditional improvement
(IECI)~\cite{gramacy2010optimization} formulation, we refer to this
proposed methodology as \emph{HW-IECI}.
We leave the systematic exploration of other acquisition functions for future work.
Note that uncertainty can be also encapsulated by replacing the indicator functions with
probabilistic Gaussian models as in~\cite{gramacy2010optimization},
whose implementation is already supported by
the used tool~\cite{snoek2012practical} and whose analysis we leave for future work.

\vspace{-7pt}
\subsection{Alternative methods supported by \emph{HyperPower}}
\vspace{-7pt}

\textbf{Constraints as Gaussian Processes -- \emph{HW-CWEI}}: We also consider the case where the constraints
are modeled by GPs~\cite{gelbart2014bayesian} using
a latent function $\hat{g} (\textbf{x}) = g_0 - g(\textbf{x})$ per constraint.
Each GP models the probability of the constraint being
satisfied $Pr(\mathcal{C}(\textbf{x})) = Pr(g(\textbf{x}) \leq g_0 )$.
In the context of our approach, to enable efficient constraint evaluation,
we evaluate the latent functions based on our models: 
$Pr(\mathcal{M}(\textbf{z}) \leq \text{MB}) \text{~~and~~} Pr(\mathcal{P}(\textbf{z}) \leq \text{PB})$.
Inspired by the Constraint Weighted EI (CWEI)~\cite{gelbart2014bayesian}
function, we refer to this second methodology as \emph{HW-CWEI}.

\textbf{Random search -- \emph{Rand}}: We consider random search as the popular
model-free alternative~\cite{bergstra2012random}. 
Once again, we exploit the insight of power modeling and early termination,
by replacing the GP-based selection with 
with random selection. We denote this method as \emph{Rand}. 

\textbf{Random walk -- \emph{Rand-Walk}}: 
Random walk methods, denoted here as \emph{Rand-Walk},
aim to ``tame'' the randomness by tuning the exploitation-exploration trade-off~\cite{smithson2016opal};
the next random point $\textbf{x}_{n+1}$ is selected around the
point $\textbf{x}^+$ with the best objective value $y^+$
over previous observations. Formally, at any step we select
from within ``neighborhood'' controlled by $\sigma^2_0$,
\emph{i.e.}, $x_{n+1} \sim  \mathcal{N} (\textbf{x}^+,\sigma^2_0)$.

In Section~\ref{sec:results}, we show that by exploiting our insights,
the \emph{HyperPower} implementations of these methods,
\emph{i.e.}, \emph{Rand}, \emph{Rand-Walk}, \emph{HW-WCEI}, and \emph{HW-IECI}, significantly
outperform their default, previously published hardware-unaware
counterparts~\cite{bergstra2012random}\cite{smithson2016opal}\cite{gelbart2014bayesian}\cite{gramacy2010optimization},
in the context of power- and memory-constrained hyper-parameter optimization.

\vspace{-5pt}
\section{Experimental Setup}
\label{sec:setup}

\vspace{-5pt}

We employ power- and memory-constrained optimization with the four discussed methods,
on two different machines, \emph{i.e.}, a server machine with
an NVIDIA GTX 1070 and a low-power embedded board NVIDIA Tegra TX1.
To train the predictive models, we profile the CNNs offline using
\texttt{Caffe}~\cite{jia2014caffe} on both the CIFAR-10 and MNIST.

We extend and implement the aforementioned four methods
on top of \texttt{Spearmint}~\cite{snoek2012practical}. 
We implement wrapper scripts around
the objective/constraint functions that are queried by \texttt{Spearmint},
that automate the generation of \texttt{Caffe} simulations, and
power/memory model evaluations. We employ hyper-parameter optimization on
variants of the AlexNet network for MNIST and CIFAR-10, with
six and thirteen hyper-parameters respectively. For the convolution
layers we vary the number of features (20-80) and the kernel size (2-5),
for the pooling layers the kernel size (1-3), and for the fully connected layers
the number of units (200-700). We also vary the learning rate (0.001-0.1),
the momentum (0.8-0.95), and the weight decay (0.0001-0.01) values.
While the considered experimental setup serves as a comprehensive basis
to evaluate \emph{HyperPower}, we are currently considering
larger networks on the state-of-the-art ImageNet dateset as part of future work.

\vspace{-5pt}
\section{Experimental Results}
\label{sec:results}
\begin{figure}[t!]
  \centering
  \vspace{-20pt}
  \includegraphics[width=0.5\columnwidth]{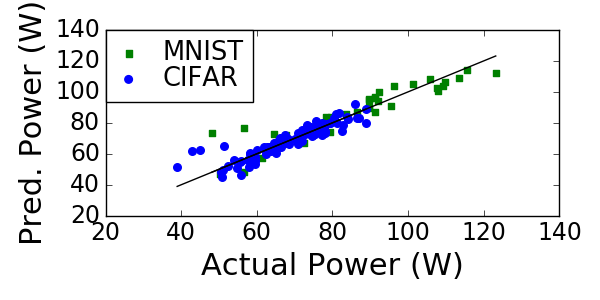}~
  \hspace*{-5pt}\includegraphics[width=0.5\columnwidth]{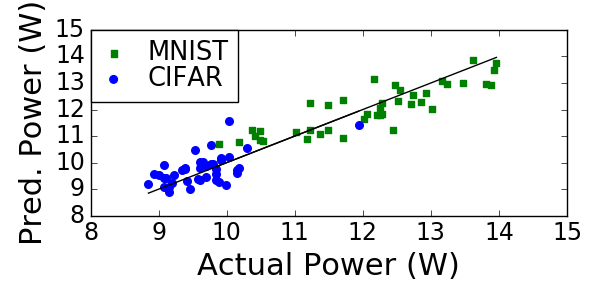}
  \vspace{-26pt}
  \caption{Actual and predicted power using our
  models for MNIST and CIFAR-10, executing on GTX 1070 (left) and
  Tegra TX1 (right).}
  \label{fig:models}
\end{figure}

\vspace{-5pt}
\textbf{Proposed power and memory models}:
First, we assess the accuracy of the power and memory models (Equations~\ref{eq:pow}-\ref{eq:mem}).
In Figure~\ref{fig:models}, we plot the predicted and actual power values, trained
on the MNIST and CIFAR-10 networks for both GTX 1070 and Tegra TX1.
Alignment across the blue line indicates good prediction results.
We observe good prediction for all tested platforms and datasets, with a
Root Mean Square Percentage Error (RMSPE) value
always less than 7\% (Table~\ref{tab:power-rmspe}) for both power and memory
models. It is worth noticing the power value ranges per device and that our proposed models
can accurately capture both the high-performance and low-power design regimes.

\textbf{Fixed number of function evaluations}:
We first assess the four methods for fixed number of
function evaluations. We apply each algorithm on the MNIST and CIFAR-10 NNs
with power constraints of 90W and 85W, respectively. As typical values
for the experiments~\cite{gelbart2014bayesian}\cite{snoek2012practical}\cite{domhan2015speeding},
we select a maximum number of 50 iterations per run (30 for MNIST);
we execute each method five times, 
and we optimize for and we report the test error results per iteration.

\begin{table}[t!]
  \vspace{-9pt}
  \caption{Root Mean Square Percentage Error (RMSPE) values of the proposed power and memory
  models.} 
  \label{tab:power-rmspe}
  \centering
  \vspace{-6pt}
  \scalebox{0.81}{
  \begin{tabular}{|c|c|c|c|c|}
    \hline
    Proposed     & MNIST   &  CIFAR-10   & MNIST   &  CIFAR-10  \\ 
    Model     &  GTX 1070  &  GTX 1070  &  Tegra TX1  &   Tegra TX1 \\ \hline \hline
    Power     & $5.70\%$  &  $5.98\%$   & $6.62\%$ &  $4.17\%$      \\ \hline
    Memory    & $4.43\%$  &  $4.67\%$   &  ~-- --~\footnote{Tegra does not support NVML API for memory measurements,
and the \texttt{tegrastats} command reports utilization and not memory consumption; for
representative comparison, we do not consider memory on Tegra.}        &  ~-- --~{\footnotesize$^\text{1}$}    \\ \hline
  \end{tabular}
  }
  \vspace{-7pt}
\end{table}

\begin{table*}
  \caption{Mean best test error (and standard deviation in parenthesis) achieved per method.}
  \label{tab:wallclock_error}
  \centering
  \vspace{-5pt}
  \scalebox{0.78}{
  \begin{tabular}{|c||c|c||c|c||c|c||c|c|}
    \hline
     ~   & \multicolumn{2}{|c||}{MNIST -- GTX 1070}  & \multicolumn{2}{|c||}{CIFAR-10 -- GTX 1070}  & \multicolumn{2}{|c||}{MNIST -- Tegra TX1}  & \multicolumn{2}{|c|}{CIFAR-10 -- Tegra TX1}    \\ \hline
    Solver       &  Default  & \textbf{HyperPower}  &  Default  & \textbf{HyperPower}  &  Default  & \textbf{HyperPower}  &  Default  & \textbf{HyperPower} \\ \hline \hline
    Rand        &   60.59\%~(41.46\%)   &  1.01\%~(0.18\%)           &  69.60\%~(28.85\%)   &  24.39\%~(3.08\%)           &   1.06\%~(0.08\%)   &  0.97\%~(0.14\%)          &   74.35\%~(22.13\%)   &  24.09\%~(1.97\%)   \\ \hline
    Rand-Walk   &   31.16\%~(41.55\%)   &  0.84\%~(0.08\%)           &  --                  &  22.88\%~(0.87\%)           &   1.04\%~(0.05\%)   &  0.90\%~(0.12\%)          &   --                  &  21.90\%~(0.59\%)   \\ \hline
    HW-CWEI     &   ~0.97\%~(~0.19\%)   &  0.85\%~(0.12\%)           &  22.09\%~(~1.01\%)   &  22.09\%~(0.35\%)           &   0.98\%~(0.13\%)   &  0.91\%~(0.07\%)          &   24.28\%~(~0.60\%)   &  22.99\%~(0.41\%)   \\ \hline
    HW-IECI     &   ~0.81\%~(~0.05\%)   &  \textbf{0.81\%}~(0.02\%)  &  22.35\%~(~1.71\%)   &  \textbf{21.81\%}~(0.38\%)  &   0.81\%~(0.02\%)   &  \textbf{0.79\%}~(0.03\%) &   23.35\%~(~1.04\%)   &  21.95\%~(0.65\%)   \\ \hline
  \end{tabular}
  }
  \vspace{-10pt}
\end{table*}

\begin{table*}
  \caption{Runtime (hours) for \emph{HyperPower} methods to reach the number of samples that their exhaustive counterparts queried.}
  \label{tab:wallclock_speedup}
  \centering
  \vspace{-5pt}
  \scalebox{0.78}{
  \begin{tabular}{|c||c|c||c||c|c||c||c|c||c||c|c||c|}
    \hline
     ~   & \multicolumn{3}{|c||}{MNIST -- GTX 1070}  & \multicolumn{3}{|c||}{CIFAR-10 -- GTX 1070}  & \multicolumn{3}{|c||}{MNIST -- Tegra TX1}  & \multicolumn{3}{|c|}{CIFAR-10 -- Tegra TX1}    \\ \hline
    Solver       &  Default  & \textbf{HyperPower}  & Speedup &  Default  & \textbf{HyperPower}  & Speedup &  Default  & \textbf{HyperPower}  & Speedup &  Default  & \textbf{HyperPower}  & Speedup\\ \hline \hline
    Rand        & 2.14  &   0.02   &  101.46$\times$ & 5.25  & 0.22  & \textbf{30.31$\times$}  & 2.08  & 0.49  &  \textbf{4.31$\times$}  & 5.35  &   0.74  &  11.78$\times$    \\ \hline
    Rand-Walk   & 2.17  &   0.02   &  \textbf{112.99}$\times$ & 5.29  & 0.46  & 17.45$\times$  & 2.14  & 1.00  &  2.15$\times$  & 5.31  &   1.13  &  \textbf{21.00$\times$}   \\ \hline
    HW-CWEI     & 2.00  &   0.30   &  ~10.22$\times$ & 5.06  & 2.46  & ~2.07$\times$  & 2.16  & 1.32  &  1.65$\times$  & 5.39  &   1.10  &  ~8.06$\times$ \\ \hline
    HW-IECI     & 2.02  &   1.81   &  ~~1.13$\times$ & 5.12  & 2.97  & ~1.74$\times$  & 2.02  & 1.65  &  1.22$\times$  & 5.22  &   1.71  &  ~3.48$\times$ \\ \hline
  \end{tabular}
  }
  \vspace{-10pt}
\end{table*}

\begin{table*}
  \caption{Increase in the number of samples that each method was able to query.}
  \label{tab:wallclock_iter}
  \centering
  \vspace{-5pt}
  \scalebox{0.78}{
  \begin{tabular}{|c||c|c||c||c|c||c||c|c||c||c|c||c|}
    \hline
     ~   & \multicolumn{3}{|c||}{MNIST -- GTX 1070}  & \multicolumn{3}{|c||}{CIFAR-10 -- GTX 1070}  & \multicolumn{3}{|c||}{MNIST -- Tegra TX1}  & \multicolumn{3}{|c|}{CIFAR-10 -- Tegra TX1}    \\ \hline
    Solver       &  Default  & \textbf{HyperPower}  & Increase &  Default  & \textbf{HyperPower}  & Increase &  Default  & \textbf{HyperPower}  & Increase &  Default  & \textbf{HyperPower}  & Increase\\ \hline \hline
    Rand        &  14.00  &  796.33    &  \textbf{57.20$\times$}  &  14.67  &   405.33   &  \textbf{27.88$\times$}  &  13.00  &   35.67    &  \textbf{2.77$\times$}  &  13.33  &  262.33    & \textbf{20.00$\times$}  \\ \hline
    Rand-Walk   &  15.00  &  316.67    &  19.16$\times$  &  13.33  &   118.33   &  ~8.86$\times$  &  14.00  &   30.67    &  2.12$\times$  &  14.33  &  ~88.67    & ~5.46$\times$  \\ \hline
    HW-CWEI     &  21.67  &  ~62.67    &  ~2.79$\times$  &  28.00  &   ~38.67   &  ~1.38$\times$  &  11.00  &   14.67    &  1.35$\times$  &  13.00  &  ~27.33    & ~1.97$\times$  \\ \hline
    HW-IECI     &  53.00  &  ~60.33    &  ~1.14$\times$  &  29.00  &   ~43.33   &  ~1.49$\times$  &  46.33  &   54.67    &  1.18$\times$  &  11.00  &  ~20.00    & ~1.75$\times$  \\ \hline
  \end{tabular}
  }
  \vspace{-10pt}
\end{table*}

\begin{table*}
  \caption{Improvement in runtime (hours) to achieve the best accuracy that the exhaustive methods did.}
  \label{tab:wallclock_benefit}
  \centering
  \vspace{-5pt}
  \scalebox{0.78}{
  \begin{tabular}{|c||c|c||c||c|c||c||c|c||c||c|c||c|}
    \hline
     ~   & \multicolumn{3}{|c||}{MNIST -- GTX 1070}  & \multicolumn{3}{|c||}{CIFAR-10 -- GTX 1070}  & \multicolumn{3}{|c||}{MNIST -- Tegra TX1}  & \multicolumn{3}{|c|}{CIFAR-10 -- Tegra TX1}    \\ \hline
    Solver       &  Default  & \textbf{HyperPower}  & Speedup &  Default  & \textbf{HyperPower}  & Speedup &  Default  & \textbf{HyperPower}  & Speedup &  Default  & \textbf{HyperPower}  & Speedup\\ \hline \hline
    Rand       &  0.50  &  0.16  & ~1.56$\times$  &  2.09    &  0.53   & \textbf{~3.97$\times$}  &  0.72    &  0.16   & ~3.64$\times$    &  2.63    &  0.48    &  4.54$\times$     \\ \hline
    Rand-Walk  &  0.69  &  0.12  & ~4.72$\times$  &  --      &  --     & --             &  1.74    &  0.27   & ~6.18$\times$    &  --    &  --    &  --    \\ \hline
    HW-CWEI    &  3.47  &  3.06  & ~6.11$\times$  &  4.12    &  2.58   & ~2.08$\times$  &  1.30    &  0.19   & ~7.39$\times$    &  5.05  &  1.24  & \textbf{4.80$\times$}  \\ \hline
    HW-IECI    &  1.61  &  0.10  & \textbf{30.12$\times$}  &  4.45    &  2.49   & ~2.13$\times$  &  1.53    &  0.14   & \textbf{11.30$\times$}    &  3.24  &  1.16  & 2.69$\times$   \\ \hline
  \end{tabular}
  }
  \vspace{-10pt}
\end{table*}

For the CIFAR-10 case, we plot the results in Figure~\ref{fig:errors}
and we make the following observations.
We confirm in Figure~\ref{fig:errors} (center) that
\emph{HW-IECI} does not select samples that violate the constraints.
This is significantly beneficial, since it allows \emph{HW-IECI} to reach the
region around the average best error in a \textbf{fifth} of
function evaluations, as shown in Figure~\ref{fig:errors} (left).
Finally, the Bayesian optimization methods outperform both random
(model-free) methods. That is, in Figure~\ref{fig:errors} (right), it is easy to
observe that most points queried by Bayesian optimization (red circles and blue squares)
are in high-performance regions, while random methods tend to select
points in low-performance regions.

\textbf{Efficient hyper-parameter optimization via power modeling and early termination
(fixed runtime)}:
Next, we evaluate the hardware-constrained hyper-parameter optimization
under maximum wall-clock runtime budget; this scenario is
important in a more commercial-standard context
when executing on a cluster~\cite{shahriari2016taking} 
and under pricing schemes in Infrastructure as a Service systems, where
speeding up the expensive function evaluation addresses not only practical but
also financial limitations related to hyper-parameter optimization~\cite{swersky2013multi,
domhan2015speeding}. We therefore repeat the exploration for three runs per method
for each considered device-dataset pair with the following constraints constraints:
85W and 1.15 for MNIST on GTX 1070, 90W and 1.25GB for CIFAR-10 on GTX 1070,
10W for MNIST on Tegra TX1, and 12W for CIFAR-10 on Tegra TX1 (no memory constraints
on Tegra{\footnotesize$^\text{1}$}). To impose upper runtime constraints, each method keeps querying
new samples as long as the total wall-clock timestamp is less than two hours
and five hours for MNIST and CIFAR-10, respectively; please note that we
allow the last sample queried right before the maximum time limit to complete
(as seen in Table~\ref{tab:wallclock_speedup}, where the
average runtime is slightly above the two and five hours spans).

\begin{figure}[t!]
  \centering
  \vspace{-17pt}
  \includegraphics[width=\columnwidth]{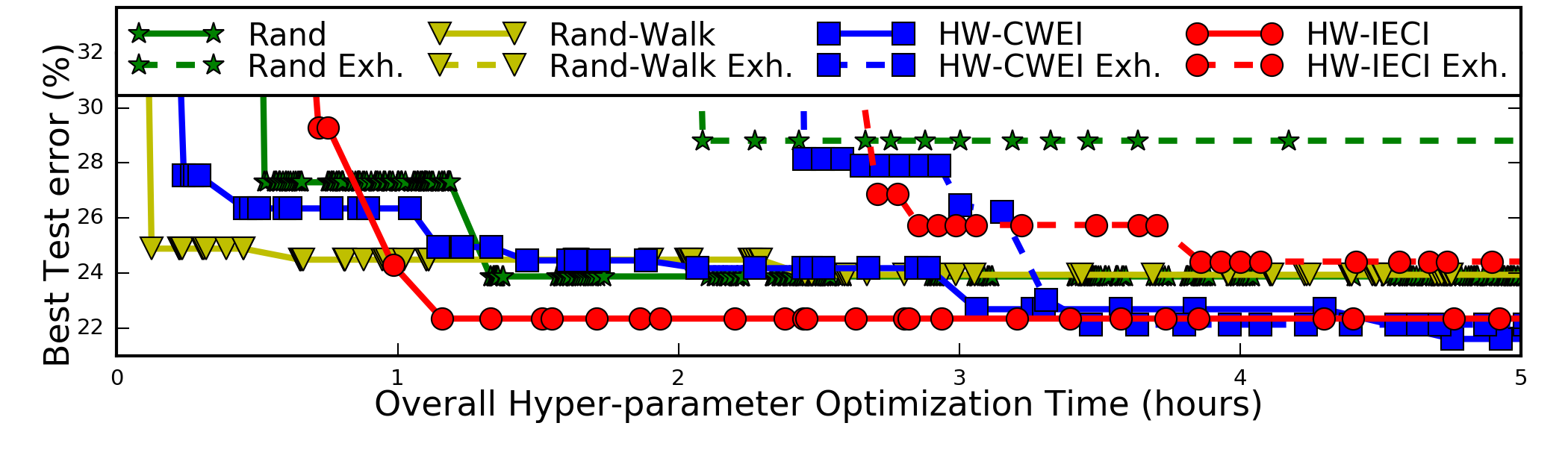}
  \vspace{-27pt}
  \caption{Capturing the benefit of using early termination and the power/memory models
  to all four considered methods; best test error on CIFAR-10 NN against the
  total hyper-parameter optimization runtime.}
  \label{fig:errors_wallclock}
  \vspace{-10pt}
\end{figure}

First, we visualize in Figure~\ref{fig:errors_wallclock} the
benefit that the power/memory models and the early termination
offer in \emph{HyperPower}. For CIFAR-10 on GTX 1070,
we repeat the 5-hour execution for each method in an exhaustive manner,
where these two enhancements are deactivated (the exhaustive versions
are shown with dotted lines in Figure~\ref{fig:errors_wallclock}). We observe that all four methods
reach a high-performance region faster that the default (exhaustive) methods, which can
be seen with all solid lines lying to the left of the dotted ones. Second,
we observe the density of the samples along the solid lines;
this is to be expected, since low-performance or violating samples can
be quickly discarded.

We present the results for all considered methods and all device-dataset pairs
in Tables~\ref{tab:wallclock_error}-\ref{tab:wallclock_benefit}, where
we compare against the constraint-unaware implementations of those methods
(these exhaustive cases are denoted as \texttt{default}, and the average
speedup values are computed as the geometric mean across all runs per case).
First, in Table~\ref{tab:wallclock_error} we report the mean and the standard deviation
of the best test error achieved by each method. As expected, the constraint-aware
versions of all four methods  supported by \emph{HyperPower}, outperform their
respective constraint-unaware counterparts, with accuracy increase by up to
$\textbf{67.6\%}$ for the case of \emph{Rand} on CIFAR-10 with Tegra TX1.

It is important to note that \emph{HyperPower} results display less variance compared to
all the constraint-unaware versions. For the random-based methods, this is because
several runs completely failed to find a high-performance region.
This is to be expected since the default exhaustive methodologies are agnostic to constraints, hence
they could keep wastefully sampling constraint-violating designs.
An extreme case of this inefficiency could be seen for both CIFAR-10 cases solved
with \emph{Rand-Walk}, which both failed to reach a feasible solution. This
highlights the key disadvantage of Random Walk methods
due to the sensitivity of their performance to the selection of the proper $\sigma_0$ value,
which defeats the purpose of automated hyper-parameter optimization altogether. These observations
for vanilla random search methods~\cite{bergstra2012random}\cite{smithson2016opal}
has significant implications, since a total of 32 hours of
server runtime was inefficiently wasted. 

Moreover, among all four versions
supported by \emph{HyperPower}, we can observe that \emph{HW-IECI} always achieves the
best results, which shows the importance of enabling \emph{a-priori} constraint evaluation through
our predictive models. Finally, we observe that \emph{HyperPower}'s enhancements allow for up to $\textbf{57.20}\times$
more function evaluations (see Table~\ref{tab:wallclock_iter}).
In terms of runtime improvement, it takes \emph{HyperPower} up $\textbf{112.99}\times$ faster
to reach the same number of function evaluations that default methods queried
(see Table~\ref{tab:wallclock_speedup}), which attests to the importance of hardware-awareness
when power/memory violating samples can be quickly discarded.
Most importantly, thanks to \emph{HyperPower}, we can reach the best test error
achieved by the exhaustive methods up to $\textbf{30.12}\times$ faster (see Table~\ref{tab:wallclock_benefit}).

\vspace{-7pt}
\section{Conclusion}
\label{sec:concl}
\vspace{-7pt}
Accounting for power and memory constraints could significantly
impede the effectiveness of traditional hyper-parameter optimization methods
to identify optimal NN configurations. In this work,
we proposed \emph{HyperPower}, a framework that enables efficient hardware-constrained
Bayesian optimization and random search. 
We showed that power consumption can be used as a low-cost, \emph{a~priori} known
constraint, and we proposed predictive models for the power and memory of NNs executing on GPUs.
Thanks to \emph{HyperPower}, we reached the number of function evaluations
and the best test error achieved by a constraint-unaware method up
to $\textbf{112.99}\times$ and $\textbf{30.12}\times$ faster, respectively,
while \textbf{never} considering invalid configurations under the proposed
\emph{HW-IECI} acquisition function.
By significantly speeding up the hyper-optimization with up
to $\textbf{57.20}\times$ more function evaluations compared to
constraint-unaware methods for a given time interval, \emph{HyperPower}
yielded significant accuracy improvements by up to $\textbf{67.6\%}$.

\section*{Acknowledgments}
\vspace{-7pt}
This research was supported in part by NSF CNS Grant No. 1564022.


\vspace{-7pt}
\bibliographystyle{IEEEtran}
\bibliography{./IEEEabrv,./dstam-date18}
%
%
%

\end{document}